\documentclass[sigconf,9pt]{acmart}
\renewcommand\footnotetextcopyrightpermission[1]{} % removes footnote with conference information in first column
\AtBeginDocument{%
  \providecommand\BibTeX{{%
    \normalfont B\kern-0.5em{\scshape i\kern-0.25em b}\kern-0.8em\TeX}}}

\setcopyright{none}
\copyrightyear{2018}
\acmYear{2018}
\acmDOI{XXXXXXX.XXXXXXX}
\usepackage{subfigure}
\usepackage{algorithm} 
\usepackage{algpseudocode} 

\acmConference[]{}{}{}
\acmPrice{15.00}
\acmISBN{978-1-4503-XXXX-X/18/06}

\begin{document}


	\author{Guangrong Zhao}
	\affiliation{%
         \country{}
	    \institution{School of Software, Shandong University}}
		\email{guangrong.zhao@sdu.edu.cn}

	\author{Yiran Shen}
		\authornote{Corresponding Author}
	\affiliation{%
  \country{}
		\institution{School of Software, Shandong University}
	}
		  \email{yiran.shen@sdu.edu.cn}
	
	\author{Ning Chen}
	\affiliation{%
  \country{}
		\institution{School of Software, Shandong University}
	}
		  \email{}
		  
	\author{ Pengfei Hu }
	\affiliation{%
  \country{}
		\institution{School of Computer Science and Technology, Shandong University}
	}
  \email{}

\author{ Lei Liu }
	\affiliation{%
  \country{}
		\institution{School of Software, Shandong University}
	}
  \email{l.liu@sdu.edu.cn}

	\author{Hongkai Wen}
	\affiliation{%
  \country{}
		\institution{Department of Computer Science, University of Warwick}
		}
			\email{hongkai.wen@warwick.ac.uk}

% %==========================================
% Abstract
% %==========================================

\begin{abstract}

Rotational speed is one of the important metrics to be measured for calibrating the electric motors in manufacturing,  monitoring engine during car repairing, faults detection on electrical appliance and etc. However, existing measurement techniques either require prohibitive hardware (e.g., high-speed camera) or are inconvenient to use in real-world application scenarios. In this paper, we propose, {\sl EV-Tach}, an event-based tachometer via efficient dynamic vision sensing on mobile devices. {\sl EV-Tach} is designed as a high-fidelity and convenient tachometer  by introducing dynamic vision sensor as a new sensing modality to capture the high-speed rotation precisely under various real-world scenarios. By designing a series of signal processing algorithms bespoke for dynamic vision sensing on mobile devices, {\sl EV-Tach} is able to extract the rotational speed accurately from the event stream produced by dynamic vision sensing on rotary targets. According to our extensive evaluations, the Relative Mean Absolute Error (RMAE) of {\sl EV-Tach} is as low as $0.3\text{\textperthousand}$ which is comparable to the state-of-the-art laser tachometer under fixed measurement mode. Moreover, {\sl EV-Tach} is robust to subtle movement of user's hand, therefore, can be used as a handheld device, where the laser tachometer fails to produce reasonable results.

\end{abstract}


\title{High Speed Rotation Estimation with Dynamic Vision Sensors}
\maketitle

% %==========================================
% Introduction
% %==========================================
%===============================================================
\section{Introduction}
\label{sec:intro}
%===============================================================

Machines and devices with rotary components are pervasive in our daily life and play significant roles in various industrial fields, such as energy, aviation, automobile and home appliance. In manufacturing, rotational speed is one of the key indicators to reflect the current working state of the machines, therefore, there is a huge demand for measuring the rotational speed with an accurate and convenient tool. In the field of appliance repairing, repairmen normally use tachometer (an instrument to measure the rotational speed) to measure the rotational speed of the electrical motors of the appliance, such as the condensing unit of air-conditioners and washing machines, to infer possible faults from the irregular rotational speed. In the automotive maintenance,  the checking of rotational speed of the wheels has become a standard item in the  annual vehicle inspection manuals~\cite{vehicle-inspections}. Real-time measurement of rotational speed is also useful for predicting the actions of flying drones because they change their flying direction and speed by adjusting the rotational speed of one or multiple propellers; the actions should be always after the change of rotational speed due to inertial effect. At last,  rotational speed calibration of some devices and equipment, such as drones, watermeters and car engines, is another typical scenario in which the rotational speed need to be calibrated precisely to ensure the devices are functioning as expected.

\begin{figure}[]
		\centering
		\includegraphics[width=\columnwidth]{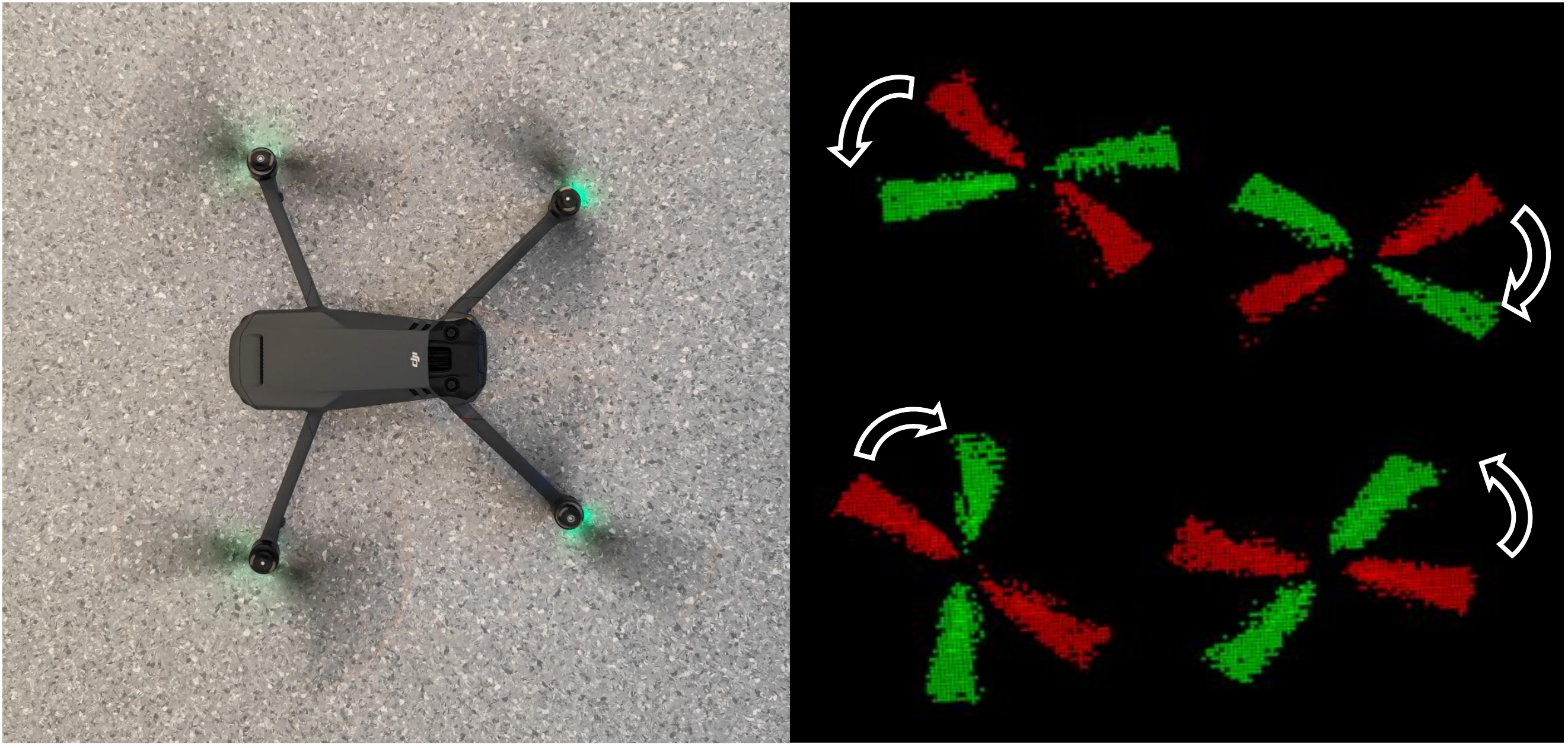}
		\caption{High-speed rotating wings of drone captured by a conventional RGB camera (left) and an event camera(right) respectively.}
		\label{fig:trigger}
\end{figure}

A number of different measurement approaches have been proposed to obtain the rotational speed of different targets under different circumstances. They are different from the requirement of physical contact (contact or non-contact) or sensing modalities (electromagnetic, laser or vision). Mechanical tachometers~\cite{Mechanical-tachometer} are a type of traditional devices to measure rotational speed of large machines via physical connection to the shaft of the targets. Electrostatic~\cite{Electrostatic}, hall-effect and optical encoder tachometers are non-contact but they must be placed in proximity to measure the rotation of the extra hardware mounted on the shafts of the targets. All the above approaches are invasive as the physical contact or extra hardware may place significant influence on the natural rotating. 
Laser tachometers~\cite{UT372,Doppler} make a step forward to more accurate and convenient measurement on rotational speed. The laser tachometer enables highly accurate (the error rate is below $0.4\text{\textperthousand}$) and low invasive measurement and can be used in reasonable working distance. Therefore, the laser tachometer has become the mainstream instrument for rotational speed measurement. However, the requirement of reflective labels on the targets still limits the application of laser tachometers under some circumstances as attaching labels may not be convenient or even impossible for some devices. Then, most importantly, although the laser tachometers are built as portable devices, it is difficult for the users to point to the extremely small label on the rotating target with the laser tachometer in hands and the accuracy in handheld mode degrades significantly according to our evaluation in Section~\ref{sec:evaluation_accuracy}. Vision-based approaches~\cite{digitalimaging,sine-varying-density,angular-velocity,Turbine-Rotor,Spectral,Electronic-Rolling,Visual-encoder} require no extra hardware on the rotating targets and can further extend the working distance with zoom lens. They also show strong environmental adaptability and robustness. However, for vision-based approaches with CCD/CMOS~\cite{digitalimaging,angular-velocity,Turbine-Rotor,Spectral,Electronic-Rolling}, the range of rotational speed to be measured is limited by the frame rate, which is normally between 30-50 fps (frames per second) and the accuracy is not good enough (error rate is over $1\%$) for high-precision measurement. As shown on the left of Figure.~\ref{fig:trigger}, high-speed rotation of drone's propellers will place significant motion blur in the recorded video of normal RGB cameras and causes failure of measurement. High-speed cameras with frame rate of few hundreds and even thousands can cover larger range of rotational speed. However, high-speed cameras are highly resource-consuming which is too prohibitive for processing on mobile devices with embedded CPUs. 

To solve the above issues, we introduce Dynamic Vision Sensing (DVS), a new sensing modality based on an emerging vision platform called event camera~\cite{Event-based-Survey}, to capture high speed rotation without motion blur. 
Figure.~\ref{fig:trigger} presents the rotating propellers of a drone landing on the floor. The left figure is an RGB frame from a video recording in a frame rate of 60 fps by an off-the-shelf smartphone camera and the right figure shows accumulative outputs of two 2 ms {\sl event streams} from DVS in red and green respectively. From the figures we can observe the rotating propellers are severely blurred in the RGB frame while the shape of the propellers are well preserved and the rotation between the two slices can be easily identified in DVS outputs. In this paper, a series of bespoke algorithms are proposed to process the DVS outputs and calculate the rotational speed accordingly and contributions can be summarized as follows:

\begin{itemize}
    \item We first introduce DVS as a new type of vision modality to mobile sensing and demonstrate its privilege in capturing high speed rotation.
    \item  An event-based tachometer, {\sl EV-Tach}, is proposed via efficient high-fidelity rotary motion sensing with DVS to estimate high-speed rotating targets. A number of algorithms are designed to process event streams for rotating objects extraction, and rotational speed estimation.
    \item We conduct extensive evaluations on the accuracy of {\sl EV-Tach} and robustness to different rotational speeds, working distances and subtle movement. According to the results, {\sl EV-Tach} achieves comparable accuracy to laser tachometer for fixed setting. The robustness of {\sl EV-Tach} to subtle movement shows it can be used in handheld mode and measure the unstable rotating targets where the laser tachometer does not work.  
\end{itemize}

The rest of the paper is organized as follows. Section~\ref{sec:related} provides related work on rotary motion measurement and DVS. Then we overview the system design in Section~\ref{sec:system-design} including the algorithms of event stream processing and rotational speed estimation. Extensive evaluations are conducted and results are presented in Section~\ref{sec:evaluation}. Finally, we discuss the advantages and limitations of {\sl EV-Tach}, and conclude the whole paper  in Section~\ref{sec:conclusion}.

% %==========================================
% Related Work
% %==========================================
%===============================================================
\section{Related Work}
\label{sec:related}
%===============================================================

In this section, we will review the work related to rotary motion sensing and dynamic vision sensor.

\noindent \textbf{Traditional Rotary Motion Sensing.}
Mechanical tachometers~\cite{Mechanical-tachometer} are physically connected to and rotate with the shaft of the target to measure the rotational speed. However, the physical contact constrains the working distance and causes inaccurate measurement due to the mass and friction of the mechanical tachometers. Electrostatic~\cite{Electrostatic} and hall-effect~\cite{hall} sensors detect the change of electromagnetic field caused by shaft-bearing fixed on the target and  the frequency of the change was estimated as the rotational speed. Optical encoder tachometers~\cite{OpticalEncoderDisc} relied on a  photoelectric sensor to detect light through the disc of encoder placed between the LED light source and photoelectric sensor. An encoder is a disc mounted on the shaft of the rotating target with opaque and transparent segments so that rotational speed can be estimated based on the pattern of the light. The electrostatic and optical encoder tachometers can be regarded as  non-contact but they must be placed in proximity to measure the rotation of extra hardware attached on the shaft of target. 
Laser tachometer~\cite{UT372,Doppler}  measures the rotational speed by detecting the small and lightweight reflective labels attached on the surface of the target.  However, the use of reflective labels may cause inconvenience during measurement especially for handheld scenarios.

\noindent \textbf{Vision-based Rotational Speed Estimation.}
There have been a number of non-contact approaches being proposed for rotational speed estimation. Wang at el.~\cite{digitalimaging} calculated the structural similarity and two-dimensional correlation between the consecutive frames, and then the similarity-related parameters were used to reconstruct a continuous and periodic signal of time-series. Fast Fourier transformation was applied to calculate the period of the signal which was used to infer the average speed of rotation. Other approach~\cite{Spectral} was also proposed to utilize the periodical change of similarity between frames and the difference was Chirp-Z transform and the parabolic interpolation based auto-correlation were applied to estimate the period in other domain.  To improve the accuracy and range of measurement, Natali at el.~\cite{Turbine-Rotor} obtained the coefficients sequence of correlation between the reference and each of the following frames. Then the rotational speed could be calculated through the short-time Fourier transform (STFT), which enabled more accurate measurement of the rotational speed of non-stationary and disturbing systems.  Instead of directly calculating the complete period of the rotation, there are some works to obtain the rotational speed by calculating the instantaneous angular speed (IAS). Zhu at el.~\cite{angular-velocity} extracted two adjacent frames from the rotational video of the objects, then the Hough transform was applied to detect straight lines, and the angular changes of these lines could be calculated. Since the interval time of the two frames was known, the angular velocity of the object could be easily obtained. However, these methods above were limited by the frame rate of the conventional RGB cameras and can only accommodate the rotational speed less than $900$rpm and the accuracy is far from our approach: the error rate is over $10\text{\textperthousand}$ which is about $25$ times worse than our proposed {\sl EV-Tach}.
In order to obtain a larger measurement range of rotational speed,  some researchers used high-speed cameras~\cite{Visual-encoder, sine-varying-density} to measure the instantaneous angular speed of rotating object. However, the cost of the high-speed cameras is prohibitive for embedded platforms and both of these methods required special-style markers attached on the rotating targets.

\noindent \textbf{Dynamic Vision Sensors.}
In this paper, we define the dynamic vision sensing as a type of vision sensing modality based on event-camera~\cite{Event-based-Survey}. The event-camera is bio-inspired and its pixels work independently to detect the change of intensity. Unlike the frame-based RGB cameras, the output of event-camera consists of nonstructural and discrete event points in spatial-temporal domain and is termed as event stream. Processing of event streams is a new topic to study. To facilitate existing methods, event streams were converted to other familiar formats, including images, graphs and 3D pointclouds. For examples, image-like representations of event streams were introduced by accumulating the event points for each pixel overtime and corresponding methods were proposed for gesture recognition~\cite{amir2017low}, gait recognition~\cite{wang2019ev} and estimating optical flow of event streams~\cite{zhu2018ev}. However, the image-like representation ignored the temporal information of event stream. Graph-based representations were proposed to preserve the spatial-temporal information of event streams. 2D-Graphs~\cite{bi2019graph} or 3D-Graphs~\cite{wang2021event} were built by selecting and connecting event points via nearest neighbor search, then graph-based convolutions were applied to extract higher-level information. The spatial-temporal event streams could also be processed as 3D pointclouds then the PointNet~\cite{qi2017pointnet} and PointNet++~\cite{PointNet++} were applied, e.g., for gesture recognition~\cite{wang2019space}. 

\noindent{\bf Most relevant work.} In~\cite{VisualFlow}, the authors proposed a method to calculate the optical flow of moving object in event stream and showed its application on estimating rotational speed of a plate with single straight line to simulate a blade. Though it also utilized the DVS as sensing modality for rotational speed estimation, the design of algorithm was not sophisticated enough to obtain accurate measurement on high-speed rotation: the method could only provide reasonable measurement for rotational speed less than $500$rpm.  Gallego at el. ~\cite{Gallego2017AccurateAV} proposed an approach to estimate the rotation of event-camera by processing the event streams. They applied contrast-maximizing edge alignment algorithm to estimate the angular velocity which was relevant to our work. However, it measured the rotation of the event-camera itself, which was fundamentally different from the goal of our work, meanwhile it produced significantly lower measurement accuracy (the error rate is around $2\%$) and sensing range (less than $167$rpm) than our approach.

% %==========================================
% Methods
% %==========================================
%===============================================================
\section{System Design}
\label{sec:system-design}
%===============================================================

\begin{figure*}[htb]
		\centering
		\includegraphics[width=\linewidth]{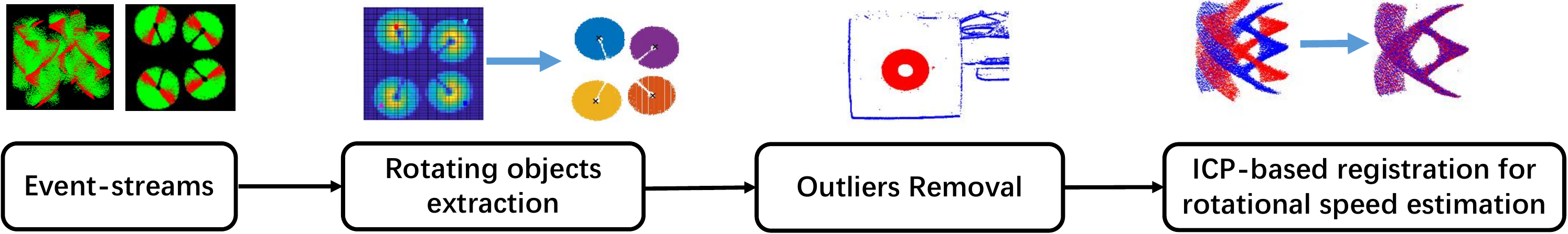}
		\caption{System overview of {\sl EV-Tach}}
		\label{fig:system}
\end{figure*}

In this paper, we propose an event-based rotational speed measurement system, {\sl EV-Tach}. The overall system design of {\sl EV-Tach} is presented in Fig.~\ref{fig:system}. It starts with recording a short-period of event stream. Then a series of event stream processing algorithms are proposed to extract multiple rotating objects from the event stream, remove outlier events to improve the data quality. At last, we propose an ICP-based registration method to estimate the transformation between two slices of event streams and calculate the rotational speed of the target.

%===============================================================
\subsection{Event Stream Processing}
\label{sec:event-processing}
%===============================================================

In this section, we will describe the event stream processing algorithms in details. The algorithms aim to extract high quality and low-dimensional event stream with single rotating target for the rotational speed estimation in next section.

\subsubsection{\bf \textit{Dynamic Vision Sensors}}
\label{sub:dvs-prem}

To make the paper self-contained, we will brief the principal background of event cameras before we describe the event stream processing algorithms in details. Dynamic vision sensors, or event cameras, are bio-inspired visual sensors developed to mimic the imaging principles of the biological retina. And in recent years, the event cameras  have been widely applied in a variety of computer vision tasks, such as super resolution, image deblurring, gesture recognition, etc. Unlike traditional RGB cameras, event cameras do not produce synchronous video frames at fixed rate, but asynchronous event streams. Specifically, pixels of the event camera work independently, to detect the change of the intensity of the scene as,
\begin{equation}
    |log~I(x,y,t_{now})-log~I(x,y,t_{previous})| < C
\end{equation}
where $I(x,y,t)$ is the intensity value of pixel $(x, y)$ at time $t$. When the change of intensity at the pixel is over the threshold $C$, an event will be released immediately. An event stream is a collection of events overtime and is represented as a stream of quadruplet $\{x,y,t,p \}$. When the event corresponds to a positive change, the polarity $p$ is $+1$ otherwise it is $-1$. 
Compared with traditional RGB cameras, event cameras posses a number of unique characteristics. As an event is launched as soon as a change is detected without global synchronization, the event streams are high in temporal resolution and low in response latency (in the order of microseconds). Event cameras save sensing energy and bandwidth as they produce events only when changes are detected. The high dynamic range (140 dB vs. 60 dB of traditional RGB cameras) enables them work greatly under challenging lighting conditions. These characteristics make event cameras have great potential for high-speed motion capture and working on resource-constrained devices.

\subsubsection{\bf \textit{Rotating Objects Extraction}}
\label{sub:dvs-detection}

One of the important merits of {\sl EV-Tach} over electromagnetic and laser tachometer is its capability on sensing multiple rotating targets simultaneously. For example, in drones manufacturing, the four independent electric-motors can be calibrated without changing the settings of the measurement. To estimate the rotational speed of multiple targets, we propose a K-means-based rotating objects extraction algorithm to isolate the events belonging to different rotating targets for further processing.

\noindent \textbf{Heatmap-based Stream-centroids Initialization: }
% Replace the centroid thing
K-means is widely used clustering algorithm in euclidean space. It literately merges the points to the nearest clusters and update the centroids accordingly until it converges. For rotary motion sensing, the resultant centroids are the locations of the rotation axes of all rotating targets. The computational complexity of K-means is low and can run in-situ on resource-constrained platforms. However, it suffers from instability and sensitivity to the initial location of centroids~\cite{initial4}: the poor choice of initial centroids may lead the algorithm fall into local optimal and result in incorrect clusters.

\begin{figure}[h]
	\centering
	\subfigure[initial centroids on heatmap]{
		\includegraphics[width=1.55in]{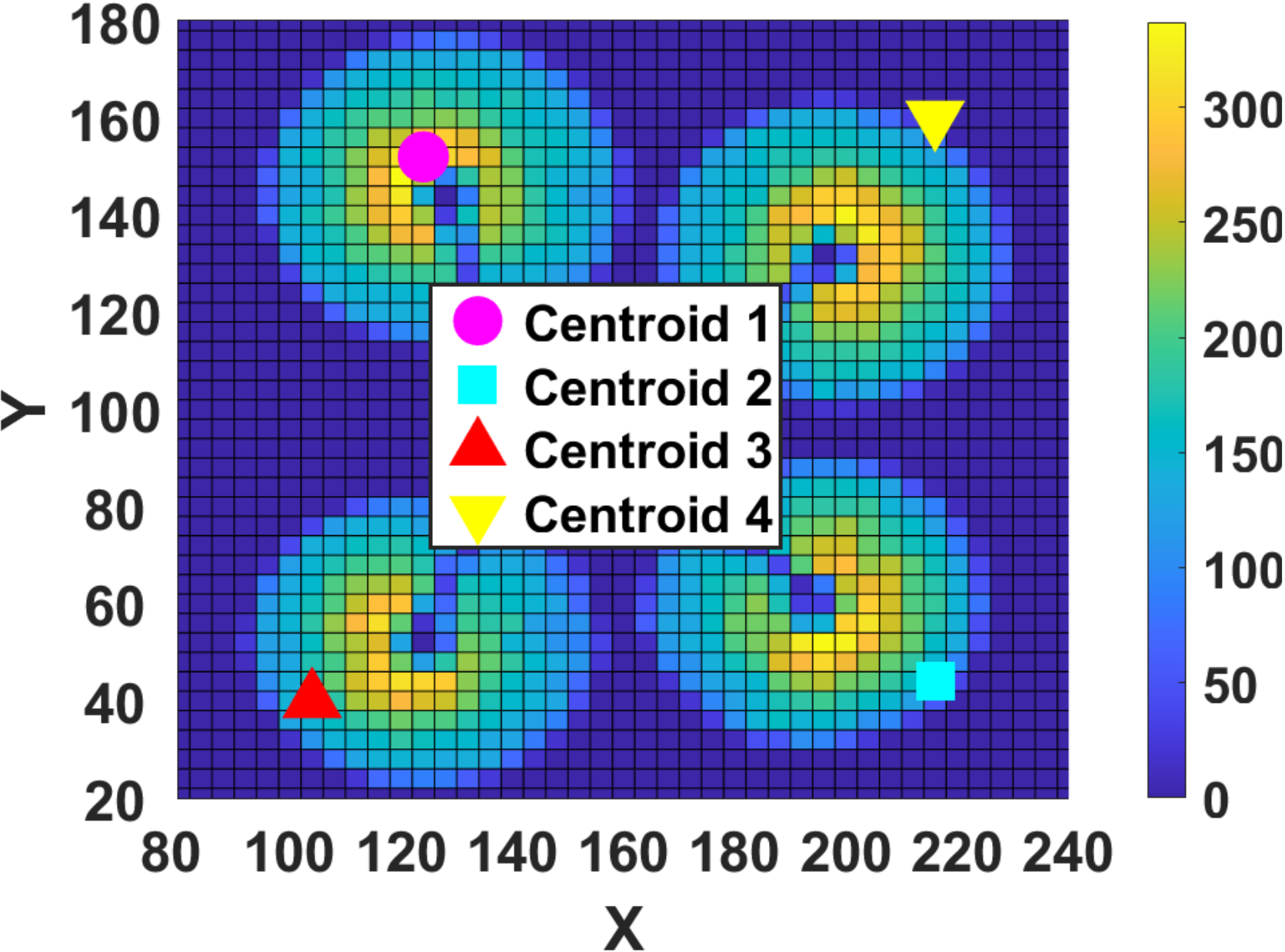}
	\label{fig:heatmap}}\hspace{0mm}
	\subfigure[clustering result]{
		\includegraphics[width=1.32in]{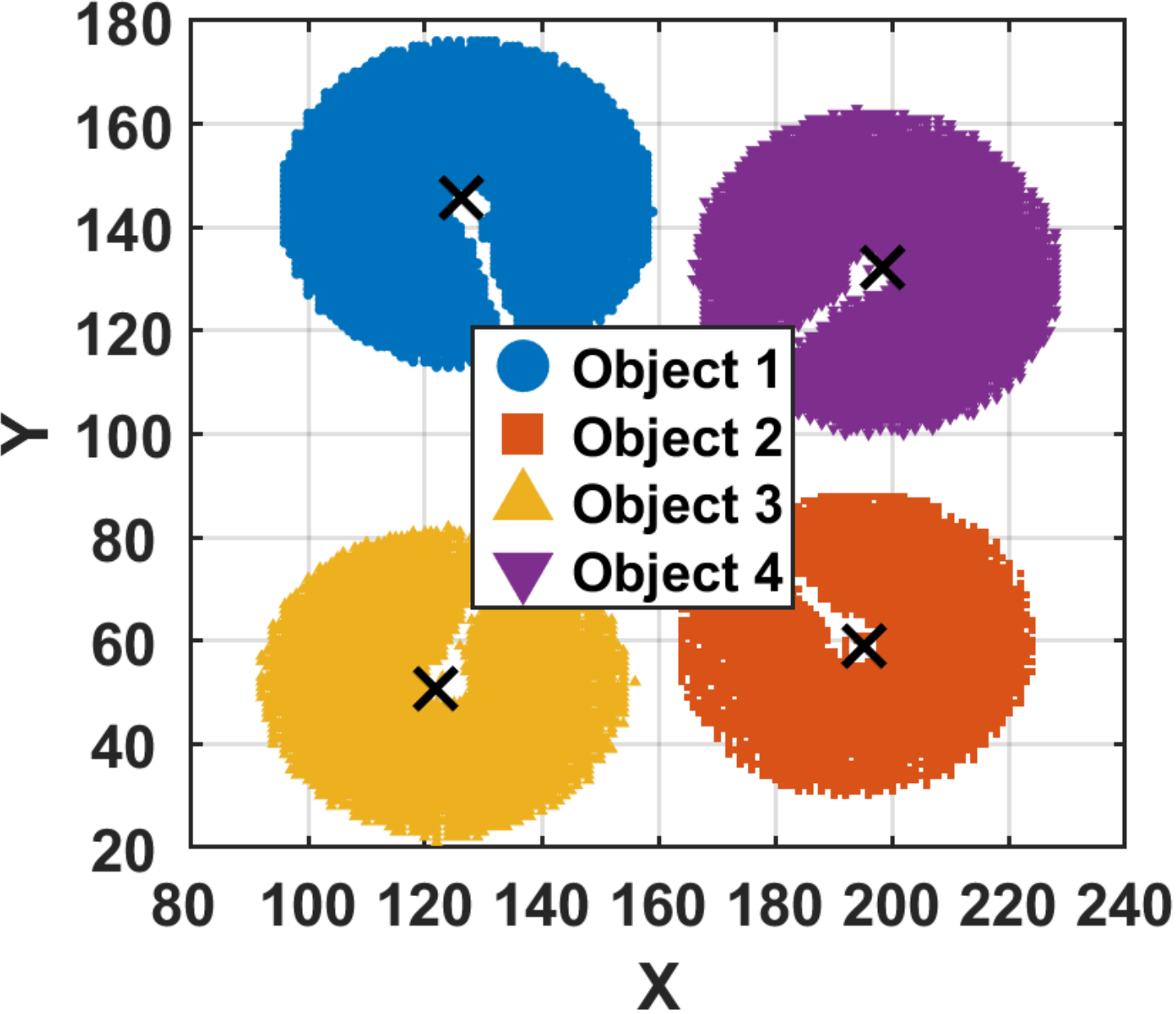}
	\label{fig:clustering_result}}\hspace{0mm}
	\caption{The initial locations of the centroids on the heatmap and the clustering results of four-propellers drone.}
	
\end{figure}

Considering the characteristics of event streams produced by rotating targets, we propose a lightweight stream-centroids initialization method based on the heatmap of accumulated events to enable reliable rotating objects extraction. After caching a fixed-length of event stream, e.g., 150ms, the number of events on each pixel are accumulated into a heatmap. The size of the grid in our setting is $4\times 4$ pixels. Figure~\ref{fig:heatmap} presents the locations of the initial centroids on the heatmap of an event stream collected from a four-motor drone. From the heatmap, we can observe, more events are generated near the center of the rotating target. To localize the initial centroids, we find the grid with the highest value (denoted as $h$) in the heatmap as the initial centroid of the first cluster. Then the remaining centroids are chosen by finding the farthest distributed grid whose value larger than $\epsilon h$ ($\epsilon=0.3$ in our evaluations and experiments), i.e., the second initial centroid is chosen as the farthest grid with over $\epsilon h$ events, the third one is chosen as the grid with largest average distance to the first two centroids and so on. Figure~\ref{fig:heatmap} shows an example of the four initial centroids chosen by our method. The principal behind this strategy is to maximize the inter-cluster distance to avoid the local optimal of K-means algorithm. Figure~\ref{fig:clustering_result} demonstrates the clustering results of the event stream generated by a four-propeller drone, where the four rotating objects are separated correctly.

\noindent \textbf{Spatial Clustering on Event Streams: }
% Replace the k means stuff
After the centroids initialization, the event stream should be segmented into multiple clusters corresponding to each individual rotating target. In each iteration, K-means-based clustering is applied to associate each event $e_i$ to one of the clusters with the nearest centroid in euclidean space and then update the centroid of each cluster by:
\begin{equation}
\hat c_{i}=\frac{1}{\left|Q_{i}\right|} \sum_{e_{j} \in Q_{i}} l(e_{j})
\end{equation}
where $\hat c_i$ is the updated centroid location,  $|Q_i|$ is the total number of events and $l(e_j)$ is the location of the event $e_j$. The events clustering and centroids updating procedures are executed alternatively until the location of the centroids remain (almost) the same, which indicates a stable clustering result has been achieved.

\noindent\textbf{Choice of $k$: } 
As the number of rotating objects in an event stream can be various, the choice of $k$ is of importance for the K-means clustering. To determine $k$ without the prior knowledge, we apply Davies-Bouldin Index (DBI)~\cite{DB} to evaluate the quality of the clustering. Specifically, we assume the candidate values of $k$ are $\{k_1, k_2, k_3..,k_i,...\}$. When $k=k_i$, a collection of $k_i$ clusters $\mathcal{Q}_{k_i}=\{Q_{1}, Q_{2},..., Q_{i},..., Q_{k_i}\}$ are obtained from K-means clustering. Then the DBIs of the clustering results with different value of $k$ are calculated and the minimal DBI is desired to maximize the inter-cluster distance and minimize the intra-cluster distance. To calculate DBI when $k=k_i$, we need first estimate the dispersion of each cluster and the separation between any of the two clusters. The dispersion of cluster $Q_{i}$ when $k=k_i$ is 
\begin{equation}
Disp(Q_{i})=\frac{1}{\left|Q_{i}\right|} \sum_{e_{j} \in Q_{i}} d\left(e_{j}, c_{i}\right)
\end{equation}
where $d(e^j,c_i)$ is the euclidean distance between any event and the centroid of the cluster. Then the separation between cluster $Q_{i}$ and $Q_{j}$ in $\mathcal{Q}$ is: 
\begin{equation}
Sep(Q_i, Q_j)=d\left(c_{i}, c_{j}\right).
\end{equation}
With the dispersion and separation, we can obtain the similarity between the two clusters  $Q_{i}$ and cluster $Q_{j}$, 
\begin{equation}
Sim(Q_i, Q_j)=\frac{Disp(Q_i)+Disp(Q_j)}{Sep(Q_i, Q_j)}. 
\end{equation}
Then the similarity between $Q_i$ and the whole collection $\mathcal{Q}$ is defined as maximum similarity between $Q_i$ and any other cluster from the collection:
\begin{equation}
SIM(Q_i, \mathcal{Q})=\max _{j=1 . . k_i, j \neq i} Sim(Q_i, Q_j)
\end{equation}
Then DBI of the clustering result when $k=k_i$ can be expressed as,
\begin{equation}
DBI_{k_i}=\frac{1}{k_i} \sum_{i=1}^{k_i} SIM(Q_i, \mathcal{Q}).
\end{equation}
Finally, the value of $k$ bringing the smallest DBI is chosen and the corresponding clustering extracts multiple rotating targets from the event stream.

\begin{figure}[h]
\includegraphics[width=0.8\columnwidth]{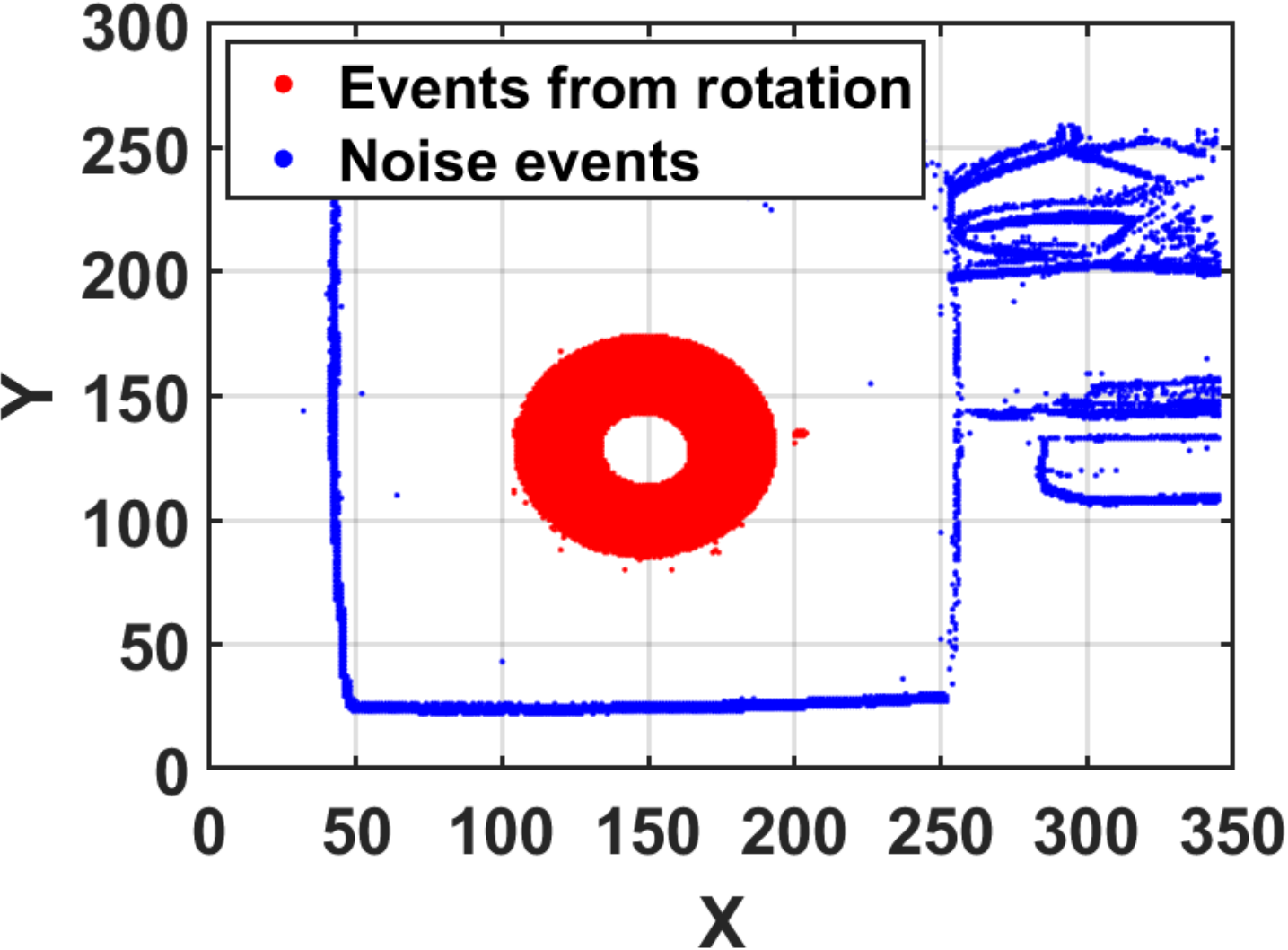}
\caption{Example of outlier events removal.}
\label{fig:outlier_remove}
\end{figure}

\noindent\textbf{Outliers Removal:}
Except for the rotating targets,  the subtle movement of users (in handheld measurement) and vibration of the host devices will also cause noticeable events in DVS and these events are regarded as outliers. Figure~\ref{fig:outlier_remove} presents the accumulated event stream in pixel domain recorded by a user holding an event-camera in front of a rotating target (the ring in red). Due to the subtle movement of user's hand, the outliers of the hosting device and edges in background (in blue) are also detected in DVS. As the rotating target causes significantly larger density of events than the subtle movement and the outliers are normally far from the centroid. Most of the valid events should concentrate around the center of rotation. To remove the outliers, we first estimate the median distance of the events to the centroid,
\begin{equation}
D_m = median\{d\left(e_1, c_i\right), d\left(e_2, c_i\right),..., d\left(e_j, c_i\right),...\}, {e^{j} \in Q_{i}}
\end{equation}
Then we set distance over three times of the $D_m$ as the threshold and the events with distance over the threshold are marked as outliers.

\noindent \textbf{Angle of Rotational Symmetry:}

For each identified rotating object, to estimate its rotational speed, we first need to track the amount of rotational motion within appropriate time frame. In many real-world applications, many rotating objects are of centrosymmetric shapes, such as propellers of drone, fans of condensing unit and wheels of automobile etc. To estimate rotational motion of these objects, we identify certain features on the rotating objects, e.g. the blades of the propellers/fans, spokes of the wheels, and track the motion of those features (i.e. angle of rotation) as a proxy for the rotational motion of the target objects. Note that for objects without those intuitive features, such as a plain rotating disk, in practice we can easily annotate them, e.g. with stickers or patterns, to create such trackable features. In addition, without loss of generality in this work we assume the rotating objects are rigid bodies, i.e., there is no significant deformation during their motion.

In this context, to accurately track the rotational motion of these features, e.g. the blades of the propellers, an important parameter to determine is the \textit{angle of rotational symmetry}, i.e. the smallest angle for which a feature can be rotated to coincide with itself or the other features. In our case, this is used to determine the appropriate length of event streams for the later ICP-based registration to avoid ambiguity. 
 
For example, Figure~\ref{fig:Rotational_motion} shows an event stream caused by a rotating object with three separated blades. 
K-means ++~\cite{kmeans++} with Davies-Bouldin Index (DBI) evaluation is applied to separate different blades in event stream with only a small number of events (e.g., $300$). Because when the number of events become large, the events generated by different blades will be entangled spatially due to the rotation. Then the angle of symmetry can be determined by the number of repeated parts, e.g., blades. 

\begin{figure}[h]
\includegraphics[width=0.8\columnwidth]{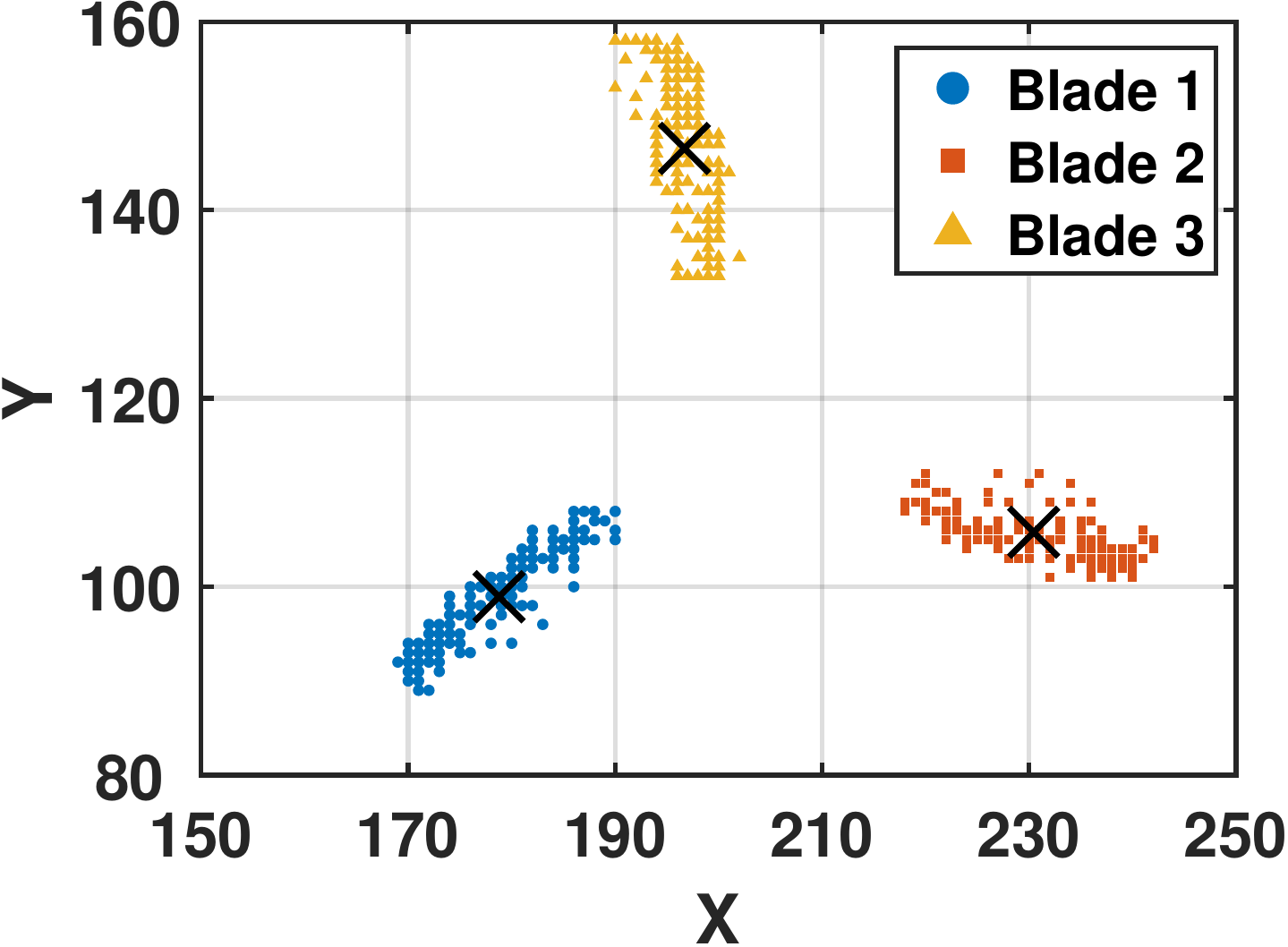}
\caption{Example of determining the angle of symmetry.}
	\label{fig:blades}
\end{figure}

%===============================================================
\subsection{Rotational Speed Estimation}
\label{sec:omega-est}
%===============================================================
% A paragraph here to given an overview of the two stages

After the event streams corresponding to different rotating targets are extracted, an ICP-based registration approach is proposed to estimate the rotational speed according to the transformation of event stream overtime. To accommodate larger range of rotational speed, the approach applies a two-stage  {\bf coarse-to-fine strategy}: initial alignment provides a coarse estimation as a feedback to the refinement stage to obtain accurate rotational speed.

\subsubsection{\bf \textit{Initial Estimation}}
\label{sub:initial-est}
\begin{figure}[h]
	\centering
	\subfigure[Before registration]{
		\includegraphics[width=1.5in]{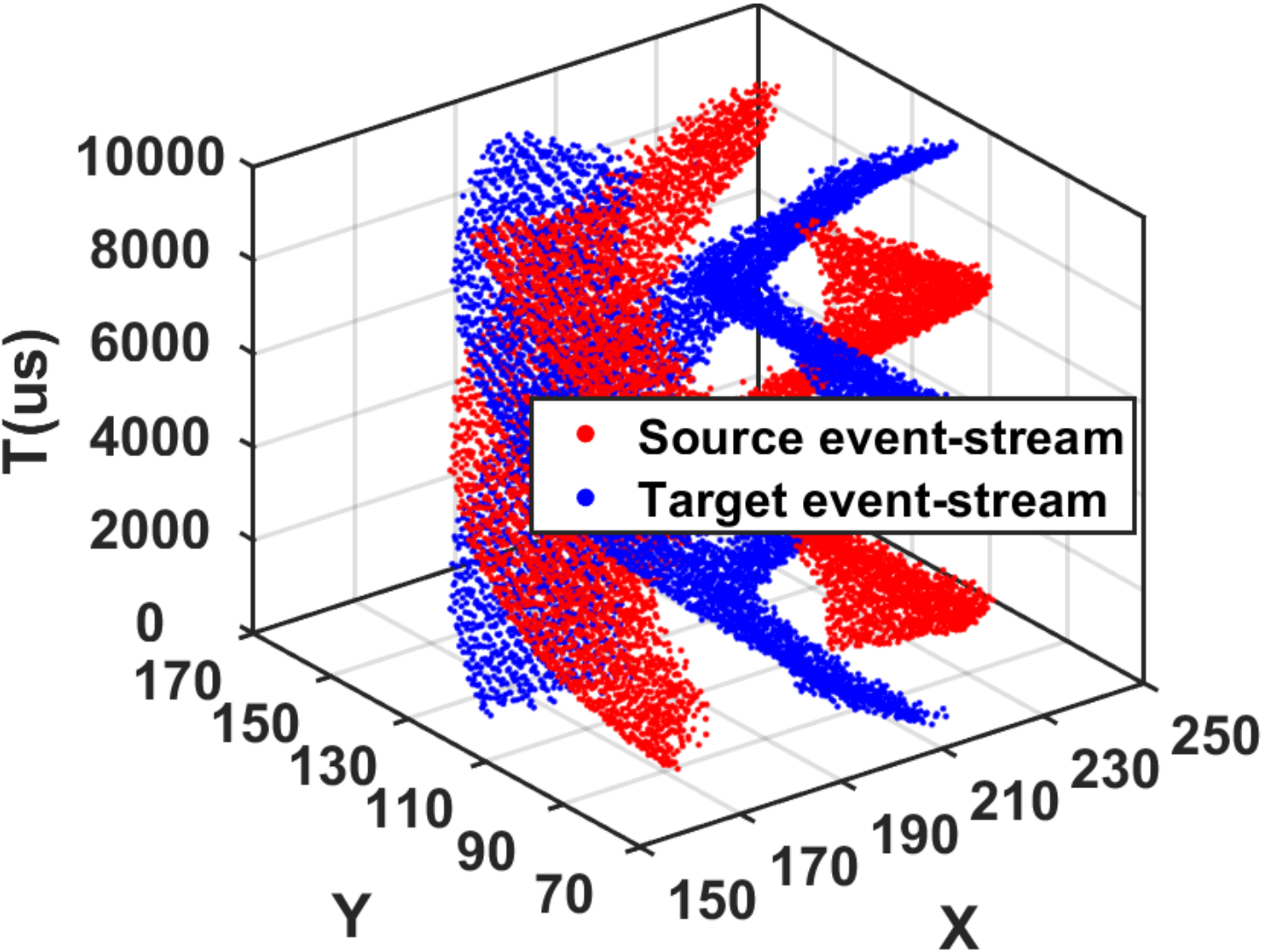}
		\label{fig:before_registration}
	}\hspace{0mm}
	\subfigure[After registration]{
		\includegraphics[width=1.5in]{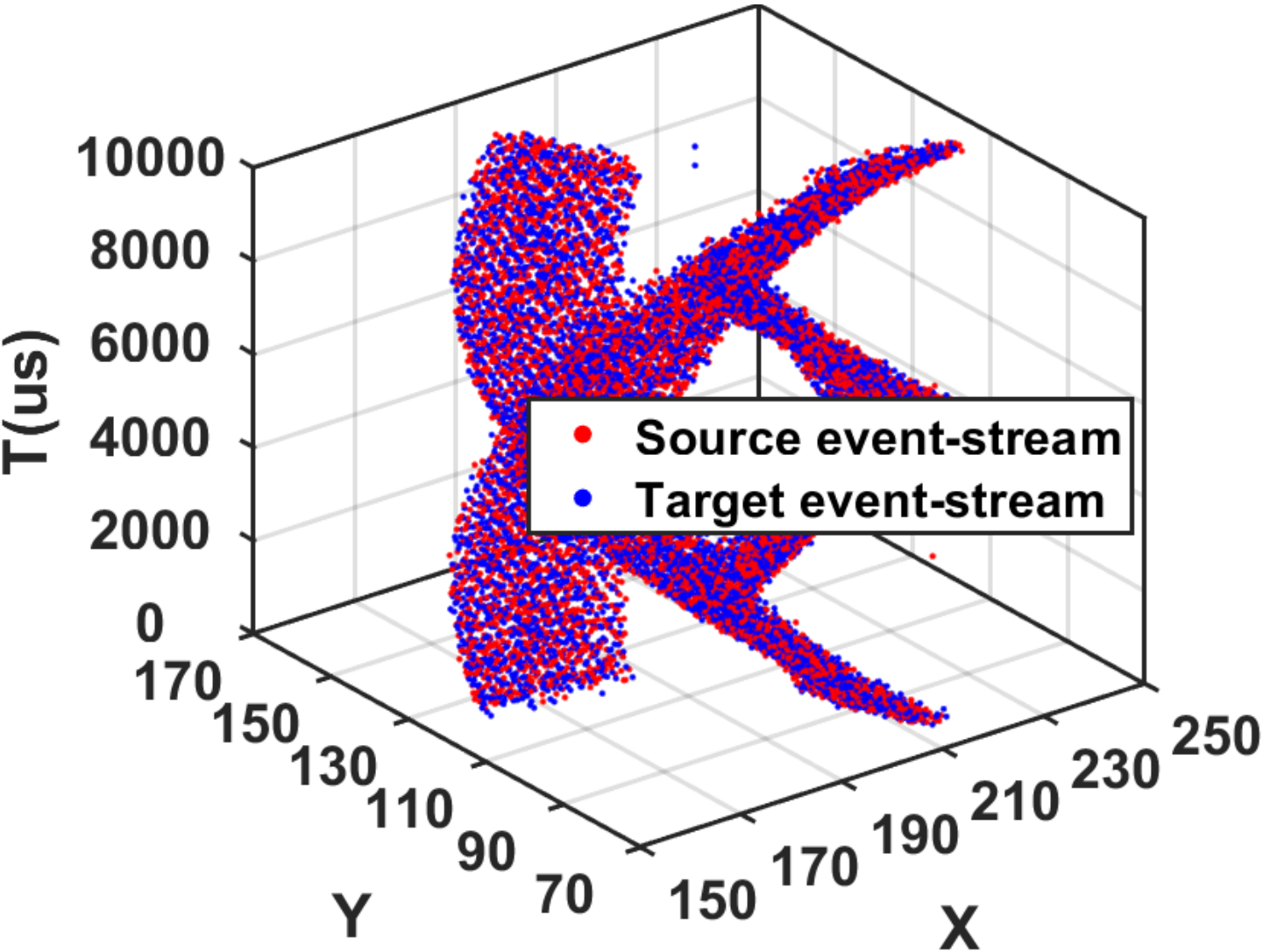}
		\label{fig:after_registration}
	}\hspace{0mm}
	\caption{Event stream registration.}
	\label{fig:Registration}
\end{figure}

In {\sl EV-Tach},  the rotational speed is calculated by estimating the angle that a propeller has rotated around its axis in a specific time. For example, Figure~\ref{fig:before_registration} presents two consecutive 10ms-slices of event stream generated by a rotating propeller with three blades. The two slices share 7ms overlap and the step between the two slices are 3ms. As the two slices of event stream are generated by the same propeller, the angle of rotation between the two slices can be obtained through aligning the two slices of event stream. In this paper, we propose an even-stream registration algorithm based on iterative closest point (ICP)~\cite{icp1,icp2,icp3}. ICP is widely used algorithm for aligning points in 3D space. For example, in pointclouds~\cite{rusu20113d} registration, It aims to find the optimal transformation (rotation and translation) from the source pointcloud to the target pointcloud by minimizing the mean square error (MSE) between the points from the source and target pointclouds after registration.

\begin{figure}[h]
		\centering
		\includegraphics[width=0.7\columnwidth]{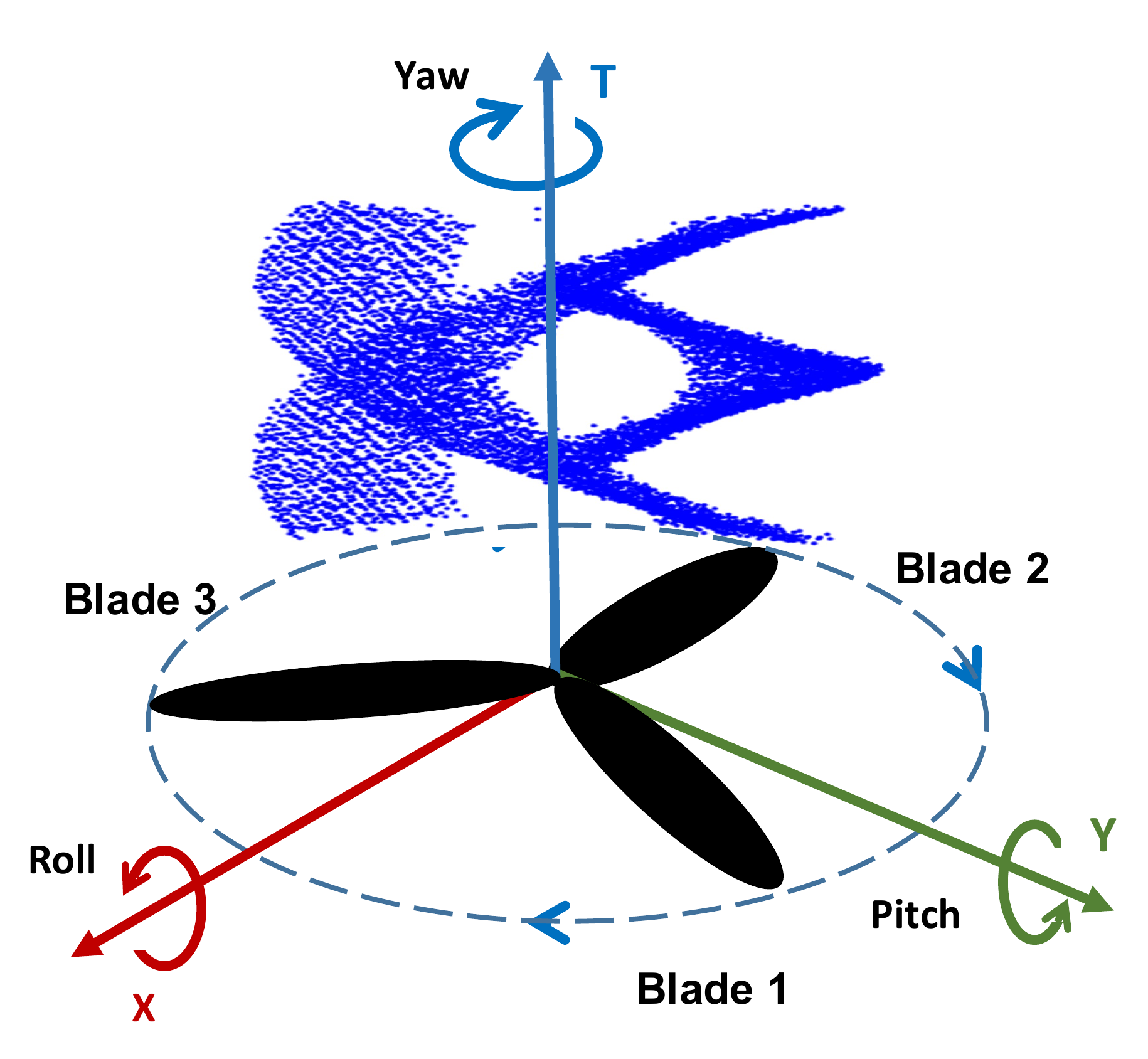}
		\caption{Demonstration of rotation in spatial-temporal space.}
		\label{fig:Rotational_motion}
\end{figure}

\noindent \textbf{Rotations in Spatio-temporal Domain:}
Before touching the details of the ICP-based registration algorithm, we first briefly introduce the rotation matrix, which is the core output we need from the event stream registration to calculate the rotational speed. The output of ICP normally consists of a translation matrix and a rotation matrix. Rotation matrix $\mathbf{R}$ describes the rotation in 3D space and can be decomposed into {\sl roll}, {\sl pitch} and {\sl yaw}, the three independent rotations around each axis according to Euler's rotation theorem~\cite{EulerAngles}.  As shown in Figure~\ref{fig:Rotational_motion}, the spatial-temporal event stream can be regarded as in three-dimension $(X, Y, T)$.  The overall rotation matrix $R$ can be decomposed into three independent rotation matrices in the spatial-temporal domain:
\begin{equation}
    \mathbf{R} = \mathbf{R}_X(\alpha)\mathbf{R}_Y(\beta)\mathbf{R}_T(\gamma)
    \label{eq:decomp}
\end{equation}
where $R_X$, $R_Y$ and $R_T$ are the rotation matrices to the three axes and $\alpha$, $\beta$ and $\gamma$ are the angles in {\sl roll}, {\sl pitch} and {\sl yaw} respectively. From Figure~\ref{fig:Rotational_motion}, we can easily identify, the {\sl Yaw} rotation around $T$-axis is directly related to the rotation of the propeller. Therefore, we only need to focus on the rotation matrix $R_T$, which can be expressed as,
\begin{equation}
\mathbf{R}_{T}(\gamma)=\left[\begin{array}{ccc}
\cos (\gamma) & \sin (\gamma) & 0 \\
-\sin (\gamma) & \cos (\gamma) & 0 \\
0 & 0 & 1
\end{array}\right].
\label{eq:trans}
\end{equation}
With ICP-based registration, we can obtain the {\sl yaw} rotation angle $\gamma$ from $R_T$. 

\noindent \textbf{ICP-based Rotational Speed Estimation: }
The ICP-based event stream registration works on two consecutive slices of event stream $\mathbf{P}$ and $\mathbf{Q}$ with length $t_l$, overlap $t_p$ and step length $t_f$ ($t_l=t_p+t_f$), where $\mathbf{P}$ is termed as source event stream and $\mathbf{Q}$ is the target event stream. ICP-based approach starts with the nearest neighbor search to find the correspondence between the events from $\mathbf{P}$ and $\mathbf{Q}$, i.e., for each event in $\mathbf{P}$, the closest event from $\mathbf{Q}$ is found in spatial-temporal domain. 
After the nearest neighbor search, we can obtain a subset $\mathbf{Q}'$ consisting events from $\mathbf{Q}$ which are nearest neighbors of events in $\mathbf{P}$. Then the co-variance between the $\mathbf{P}$ and $\mathbf{Q}'$ is,
\begin{equation}
cov =\sum_{i=1}^{N}\left(p_{i}-\bar{p}\right)\left(q'_{i}-\bar{q}'\right)^{T}
\end{equation}
where $N$ denotes the number of events in $\mathbf{P}$. $p_i$ and $q'_i$ are the spatial-temporal positions of the $i_{th}$ event from $\mathbf{P}$ and $\mathbf{Q}'$ respectively. $\bar{p}$ and $\bar{q}'$ are the spatial-temporal positions of the centroids. Then supportive vector decomposition (SVD) is applied to factorizing the co-variance matrix, i.e.,
\begin{equation}
    cov = U\Sigma V^T.
\end{equation}
The rotation matrix is $\mathbf{R}=VU^T$ and translation matrix is $\mathbf{T}_r=\bar{q}'-\mathbf{R}\bar{p}$. Along with Eq~\eqref{eq:decomp} and Eq~\eqref{eq:trans}, the yaw rotation angle $\gamma$ can be estimated and the source event stream is transformed according to the rotation and translation matrices. Then the operations above are repeated and $\gamma$ from each iteration is accumulated:
\begin{equation}
    \gamma_{acc}=\gamma_{acc}+\gamma
\end{equation}
The iteration terminates when yaw rotation diminishes, i.e., $\gamma/\gamma_{acc}<0.001$. 

It is worth noting that, DVS is noisy and non-structural, the source and target event streams normally cannot perfectly aligned. To reduce the influence of misalignment on the estimation of rotation, we apply bi-directional registration by simply switching the source and target event streams. The average yaw rotation $\bar{\gamma}$ is adopted to calculate the rotational speed (in rpm) from initial alignment:
\begin{equation}
r_{init} = \frac{\bar{\gamma}}{2\pi t_s}\times 60.
\label{eq:rpm}
\end{equation}

\subsubsection{\bf \textit{Estimation Refinement}}
\label{sub:est-refinement}
For the initial alignment stage, there is no prior knowledge on how fast the rotational speed is. Therefore, to accommodate high rotational speed, we choose a small step length, i.e., $t_s = 1ms$, in case the event streams from different blades are overlapped and lead to ambiguity in rotation estimation.  However, when the rotational speed is slow, e.g., less than 1000rpm, only very few events are generated by the rotating object within the super short period. Considering the noisy nature of DVS, the estimation from initial alignment can be unreliable. To improve the accuracy of estimation, we propose a simple but effective refinement approach based on the coarse result from initial alignment. According to the rotational speed $r_{init}$ from initial alignment, we can extend $t_s$ to include more events meanwhile avoid the ambiguity caused by central symmetry: $t_s$ cannot lead to rotation over the angle of symmetry mentioned above, otherwise, the ICP-based registration will align two different parts of the object together and causes incorrect estimation on rotation. According to above constraints, the new step length can be inferred as,
\begin{equation}
t'_{s} = \eta \frac{60}{2r_{init}}\cdot\frac{\theta_c}{2\pi}
\end{equation}
where $\theta_c$ is angle of central symmetry determined above, $\eta$ is an scale factor (<1) to accommodate the inaccuracy of initial alignment. Then the new step length $t'_{s}$ is adopted to run the ICP-based event stream registration again to obtain a refined estimation on rotational speed.
%===============================================================

% System Design
% Event Stream Processing
% Event-based Rotational Speed Estimation

% %==========================================
% Evaluation
% %==========================================
\section{evaluation}
\label{sec:evaluation}

In this section, we evaluate our proposed {\sl EV-Tach} on datasets collected from monitoring the rotation of a customized device and compare it with laser tachometer on accuracy, robustness, convenience, etc. 

\subsection{Evaluation Setup}
\begin{figure}[h]
	\centering
	\subfigure[DAVIS346 and UT372]{
		\includegraphics[width=1.55in]{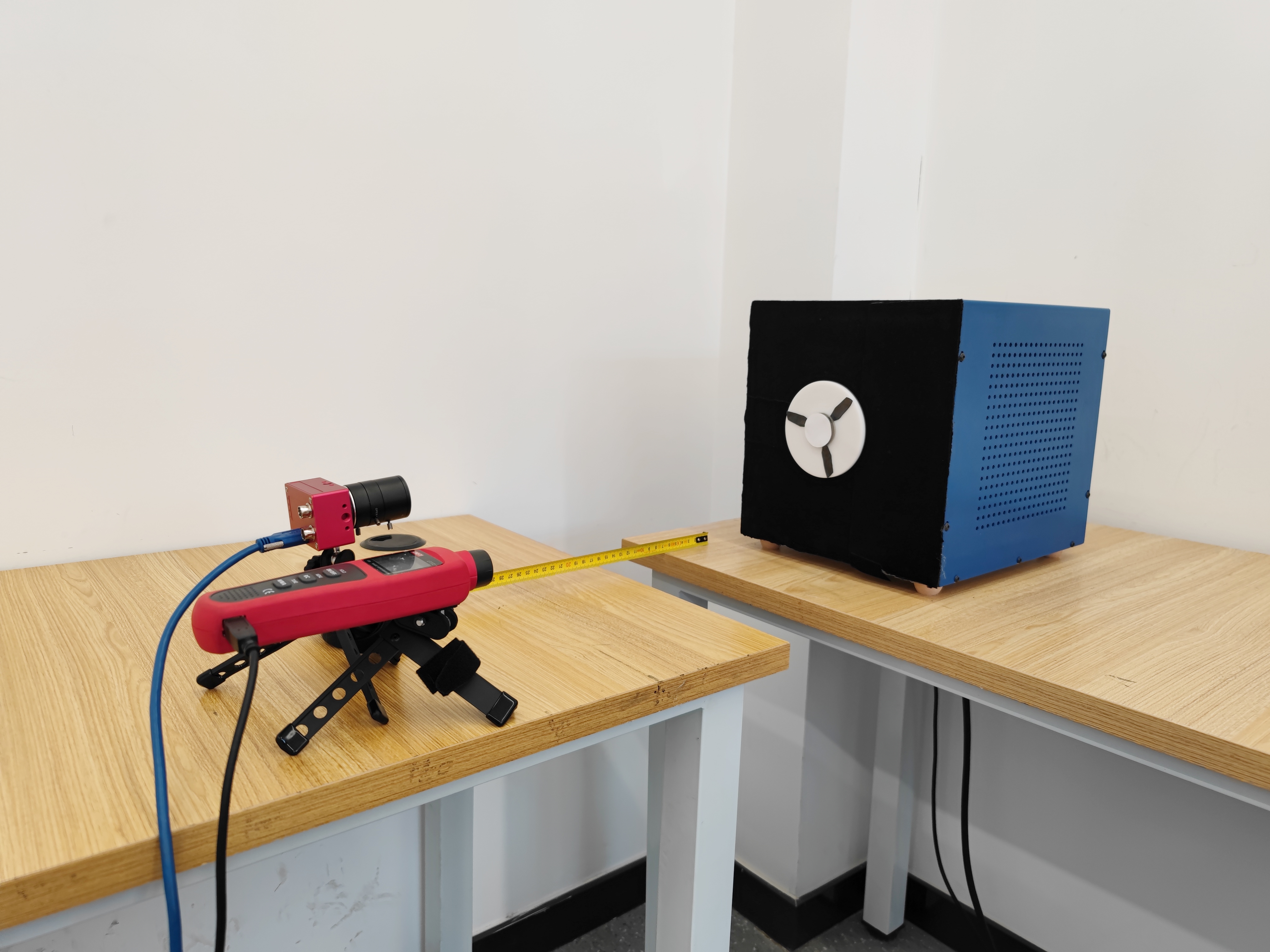}
	}\hspace{-1mm}
	\subfigure[DAVIS346 handheld]{
		\includegraphics[width=1.55in]{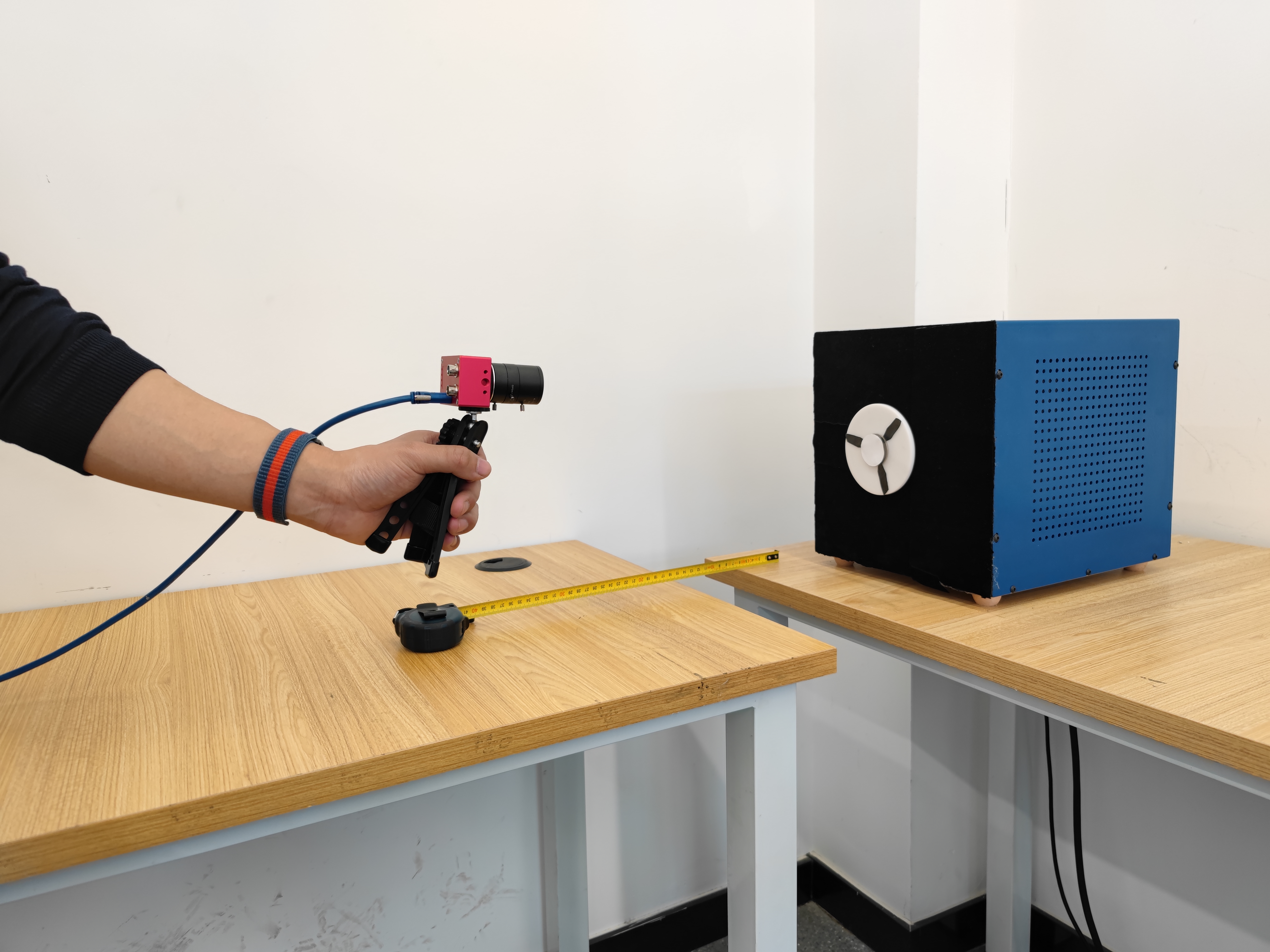}
	}\hspace{0mm}
	\caption{Experiment setup for data collection.}
	\label{fig:davis laser}
\end{figure}

\noindent{\bf Data Collection: } 
As shown in Figure~\ref{fig:davis laser}, during data collection, an event-camera is used to collect raw event streams of a rotating target on the customized device and laser tachometer is also deployed as benchmark. The customized device is equipped with a servo motor whose rotational speed can be precisely controlled through an interface on a laptop and the highest rotational speed is $6000$rpm. A white plate is connected to the motor shaft as the rotating target and ``propellers'' can be printed and attached on the plate as requirement. The event-camera is DAVIS346~\cite{davis346} whose spatial resolution is $346\times 260$ and temporal resolution is $20\mu s$. DAVIS346 comes with a vari-focal CS-mount lens which can be used to extend the measurement distance. The laser tachometer (UNI-T UT372) provides high precision measurement with relative error of $\pm 0.4\text{\textperthousand}$ and the results can be easily streamed to a computer via cable. Moreover, the measurement distance ranges from $20-50cm$. The datasets are collected by changing the rotational speed of the servo motor, distance to the target, different number of blades, and etc.

\noindent{\bf Evaluation Metrics:}
Relative mean absolute error (RMAE) is adopted to present the accuracy of measurement and defined as,
\begin{equation}
RMAE=\frac{1}{M}\sum_{i=1}^M\frac{|r_{i}-r_{gt}|}{r_{gt}} 
\end{equation}
where $M$ is the number of tests and $M$ = 30 in the following evaluation; $r_i$ is the $i_{th}$ measured rotational speed and $r_{gt}$ is the ground-truth. Low RMAE means high measurement accuracy. 

\noindent{\bf Competing Methods:} 
In the evaluations on accuracy, we consider four different measurement methods to compare the accuracy of {\sl EV-Tach} to the state-of-the-art laser tachometer, which are:

\begin{itemize}
  \item {\bf DVS-Fixed}: DAVIS346 is fixed on a tripod and placed on the table while recording event streams via DVS.
   \item {\bf DVS-Handheld}:DAVIS346 is held in hand of a user while recording the event streams. 
    \item {\bf Laser-Fixed}: laser tachometer is fixed on a tripod and placed on the table. It points to a $1cm\times 1cm$ reflective label on one of the blades. 
     \item {\bf Laser-Handheld}: laser tachometer is held in hand of a user.  The user tries to point to the reflective label while measuring the rotation. 
\end{itemize}

\subsection{Evaluation on Accuracy of Rotational Speed Estimation}
\label{sec:evaluation_accuracy}

In this section, we will provide extensive evaluations on the accuracy of {\sl EV-Tach} against different parameter settings, including rotational speed, measurement distance, number of blades and host vibration. Laser tachometer, as the state-of-the-art rotational speed measurement tool, is chosen as the benchmark to compare with {\sl EV-Tach}.

\subsubsection{\bf \textit{Evaluation on Coarse-to-fine Alignment}} 
\label{sec:evaluation_refinement}
\begin{figure}[h]
		\centering
		\includegraphics[width=0.8\columnwidth]{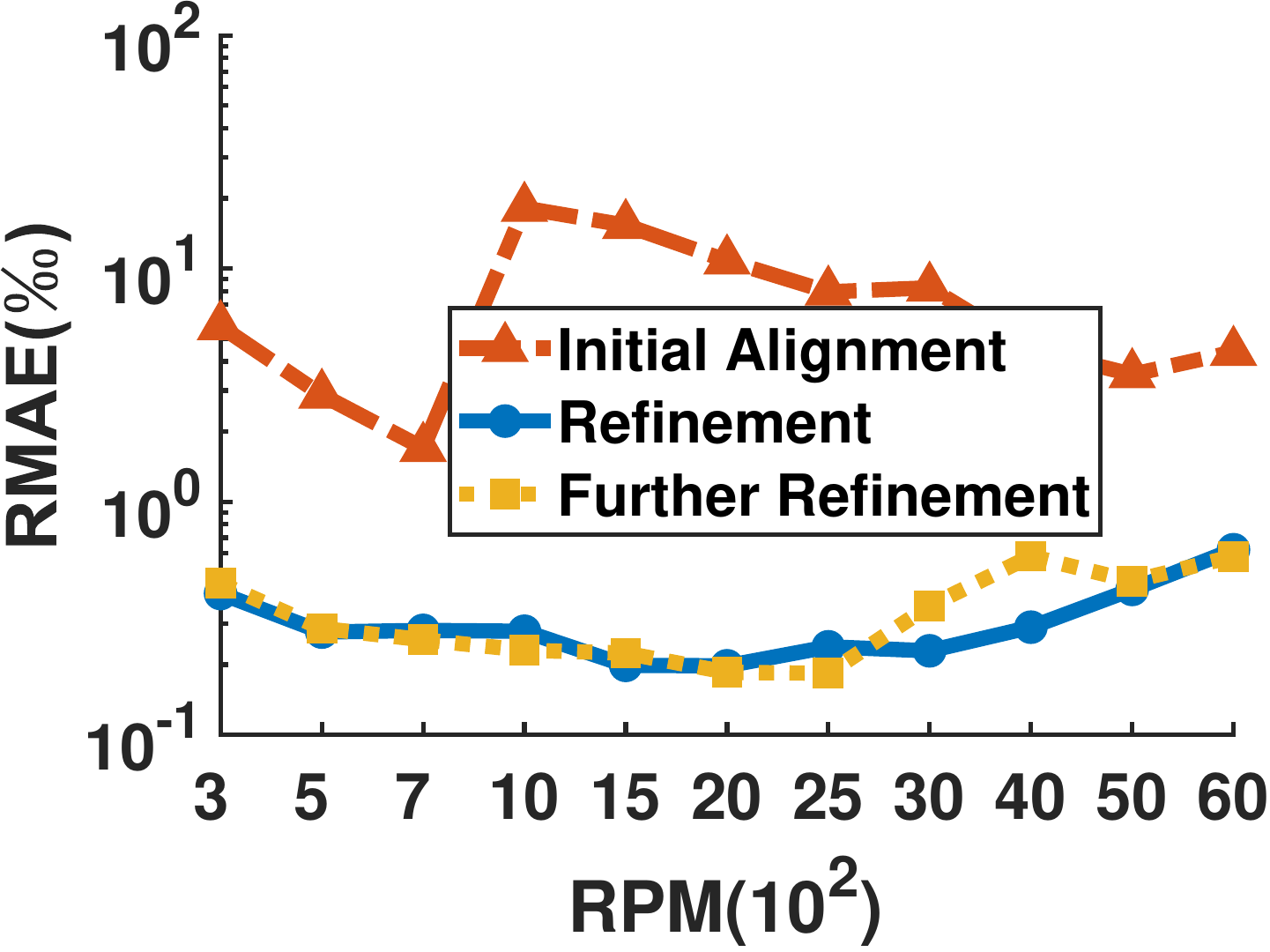}
		\caption{RMAE from different components of event stream registration.}
		\label{fig:Initial_further}
\end{figure}

{\sl EV-Tach} applies a coarse-to-fine strategy to refine the estimation obtained from initial alignment. To show the refinement stage really works and no further refinement is needed, we compute the RMAE of the outputs of the three different components of {\sl EV-Tach} ({\bf DVS-Fixed}) including initial alignment, refinement and further refinement. The RMAEs against different rotational speeds, from $300$rpm to $6000$rpm, are shown in Figure~\ref{fig:Initial_further}. By comparing different stages, we can observe, the first refinement effectively reduces the RMAE for all rotational speeds compared with initial alignment. For example, via refinement, the average of RMAEs across all rotational speeds drops significantly from $0.76\%$ to $0.031\%$, which means approximately 25 times improvement on accuracy. Moreover, further refinement cannot guarantee noticeable improvement and consumes extra resources, therefore, a two-stage strategy with initial alignment and one-time refinement is sufficient to obtain an accurate measurement.

\subsubsection{\bf \textit{Evaluation on Different Rotational Speeds}}
\label{sec:evaluation_speed}

\begin{figure}[h]
		\centering
		\includegraphics[width=0.8\columnwidth]{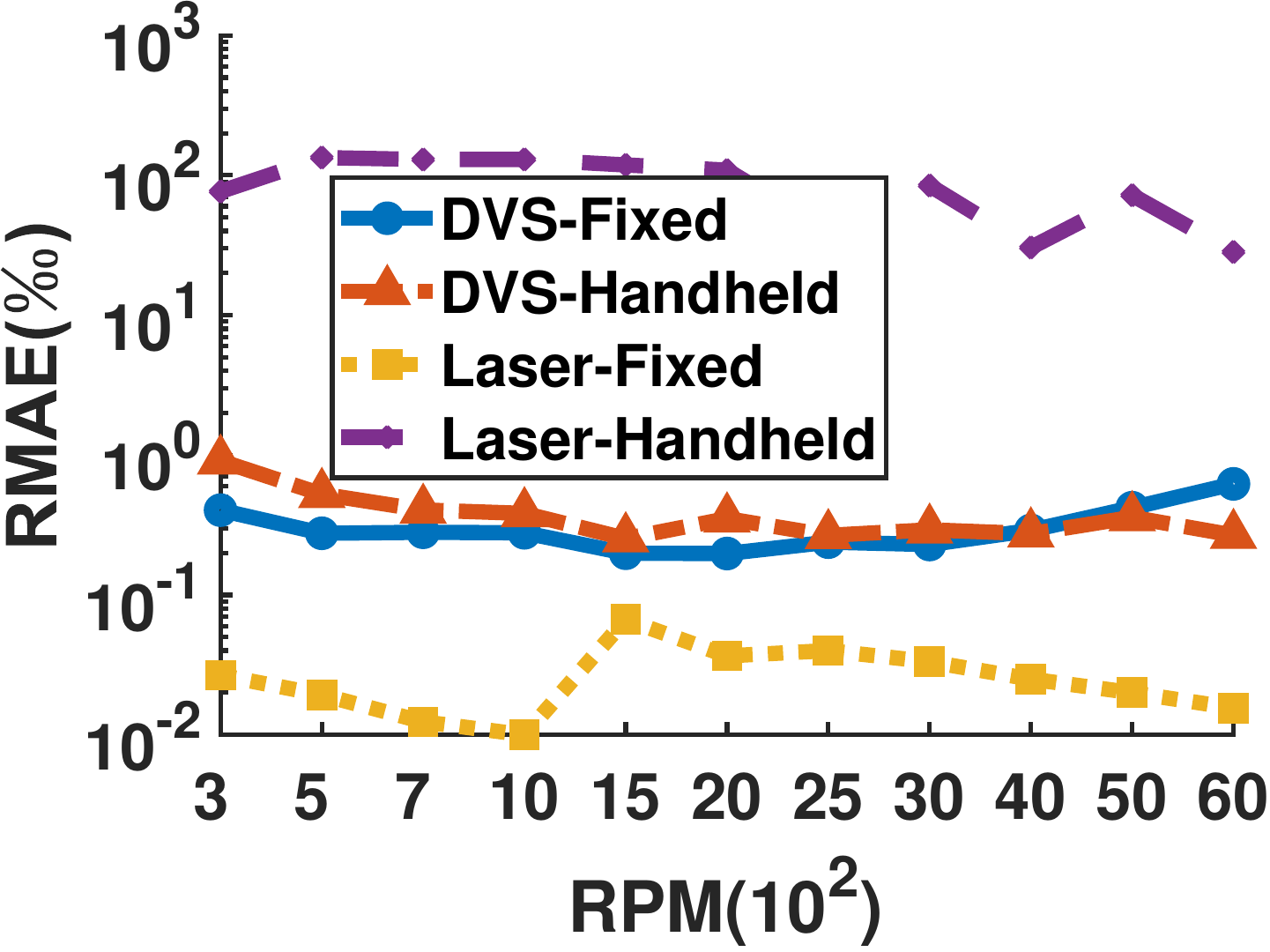}
\caption{RMAE of different measurement methods on different rotational speeds.}
\label{fig:rpm}

\end{figure}

From this section, we will compare {\sl EV-Tach} with the laser tachometer under different circumstances. First, the four different measurement methods {\bf DVS-Fixed}, {\bf DVS-Handheld}, {\bf Laser-Fixed} and {\bf Laser-Handheld} are evaluated against different rotational speeds. The measurement distance is set as $40$cm. By changing the rotational speed of the servo motor from $300$rpm to $6000$rpm, we can obtain the corresponding RMAEs of the four different methods as shown in Figure~\ref{fig:rpm}. As mentioned before, each RMAE is obtained from averaging the results of $30$ independent tests and five users are recruited for the handheld measurement. From the figure, we can observe, three methods, except for {\bf Laser-Handheld}, can produce accurate measurement with RMAEs less than $1\text{\textperthousand}$.
Especially, {\bf Laser-Fixed} achieves the lowest RMAE ($<0.1\text{\textperthousand}$). However, when the laser tachometer is held in hand, the average of RMAEs of {\bf Laser-Handheld} rockets to  $87\text{\textperthousand}$ due to the subtle movement of user's hand. When it comes to {\sl EV-Tach}, the {\bf DVS-Fixed} and {\bf DVS-Handheld} methods produce similar RMAEs and the average of RMAEs of {\bf DVS-Handheld} is below $0.4\text{\textperthousand}$ which is over 210 times better than {\bf Laser-Handheld}. Therefore, we can claim that, {\sl EV-Tach} is robust to the subtle movement. While the laser tachometer, though designed as a portable device, is not suitable for handheld measurement. It is worth noting that, it is very hard to point to the small reflective label attached on the blade when the target is fast rotating. It normally takes at least tens of seconds for user to point to the correct spot then it deviates easily from the label due to subtle movement of hand. Comparatively, {\sl EV-Tach} is significantly more convenient. Users only need to make the camera approximately face to the front of the rotating target and the procedure is in no time. According to our observation, 20-degrees deviation from the front view is allowed. Therefore, {\sl EV-Tach} is superior over laser tachometer for ease of use.

\subsubsection{\bf \textit{Evaluation on Measurement Distance}}
\label{sec:evaluation_distance}

\begin{figure}[h]
		\includegraphics[width=0.8\columnwidth]{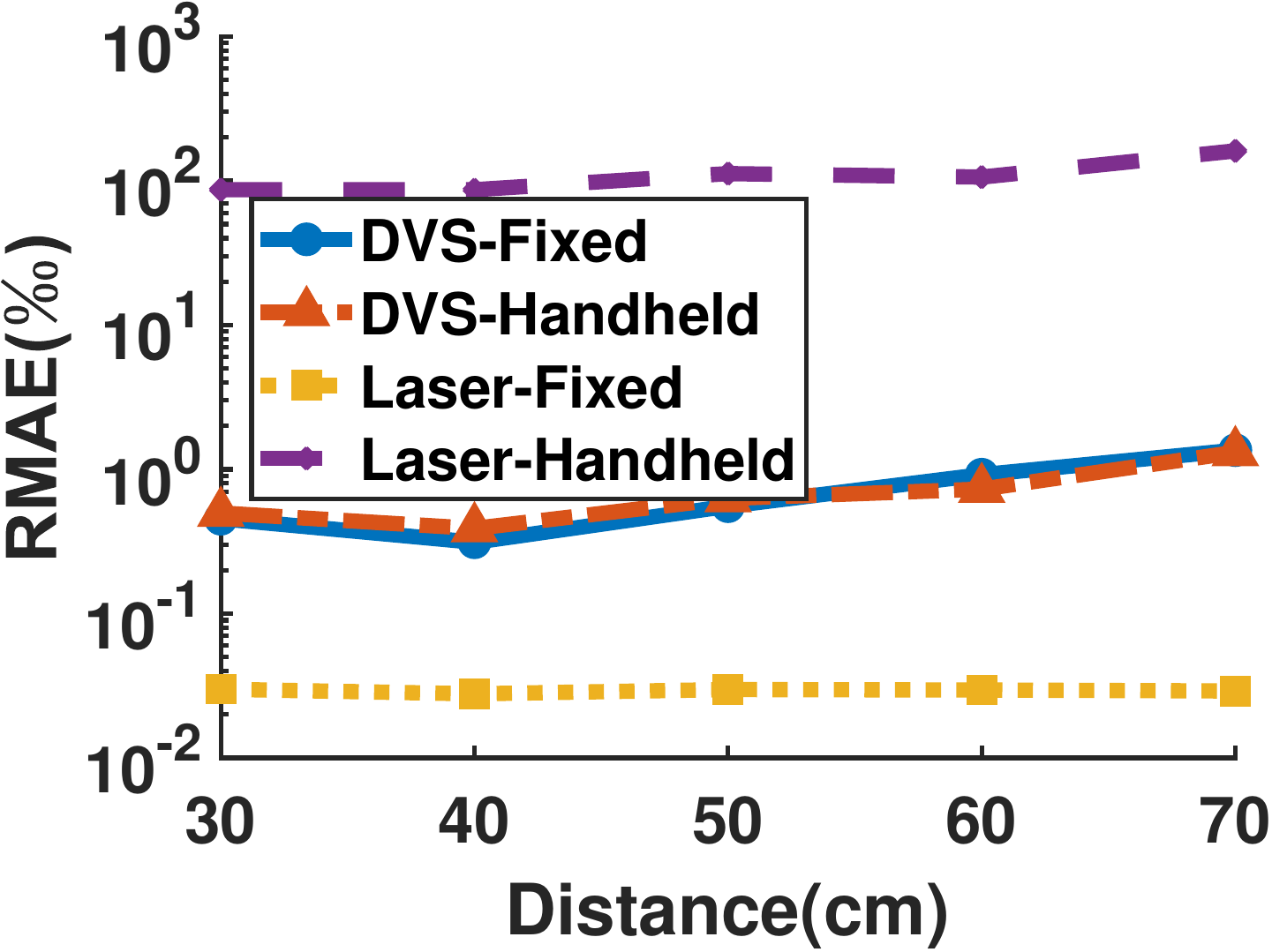}
		\caption{RMAE of different measurement methods on different measurement distances.}
		\label{fig:distance}
		\end{figure}

The distance to the rotating target during measurement can be various in use. In this section, we evaluate the accuracy of the four methods against different measurement distances. Again, RMAEs of the four methods are computed by gradually increasing the measurement distance from $30$cm to $70$cm and the results are shown in Figure~\ref{fig:distance}. Each RMAE in the figure is obtained from averaging the RMAEs obtained under different rotational speeds. From the results we can observe, {\bf Laser-Fixed} is not affected by the measurement distance: as far as the reflection of laser can reach, it will produce stable and accuracy measurement. 
The RMAEs of the remaining methods all increases with the growth of measurement distance due to different reasons. For {\bf Laser-Handheld}, with the increase of distance, it becomes more difficult for users to point the laser to the small reflective label and keep not deviate from the correct spot during the measurement.  
The accuracy of {\sl EV-Tach} approaches, {\bf DVS-Fixed} and {\bf DVS-Handheld}, declines as the event stream shrinks with the growth of distance. However, it can be solved by using a zoom lens on event-camera. For example, By changing the focal-length of DAVIS346 from $8$mm to $4$mm, we can zoom-in on the rotating target. According to our evaluation, when the measurement distance is $100$cm, the average of RMAEs of {\bf DVS-Handheld} with $4$mm focal-length is below $0.78\text{\textperthousand}$ which is similar to that of {\bf DVS-Handheld} with $8$mm in $50-60$cm measurement distance. 

\subsubsection{\bf \textit{Evaluation on Different Number of Blades}}

Generally, as three-blade propellers are the most common rotating targets to be seen~\cite{threeblades}, in the evaluations above, we set the number of blades to be three. However, it is possible the rotating targets are various in number of blades. For example, most of drones are equipped with two-blade propellers. We evaluate the accuracy of  {\sl EV-Tach} on estimating the rotational speed of propellers with two, three and four blades respectively. By gradually changing the rotational speed of servo motor, the RMAEs are calculated and presented on Figure.~\ref{fig:blades}. From the results we can observe, {\sl EV-Tach} achieves similar accuracy of measurement for all types of blades and the average of RMAEs are $0.32\text\textperthousand$, $0.31\text\textperthousand$ and $0.39\text\textperthousand$. Therefore, {\sl EV-Tach} can work on the rotating propellers with different number of blades.

\begin{figure}[h]
	\includegraphics[width=0.8\columnwidth]{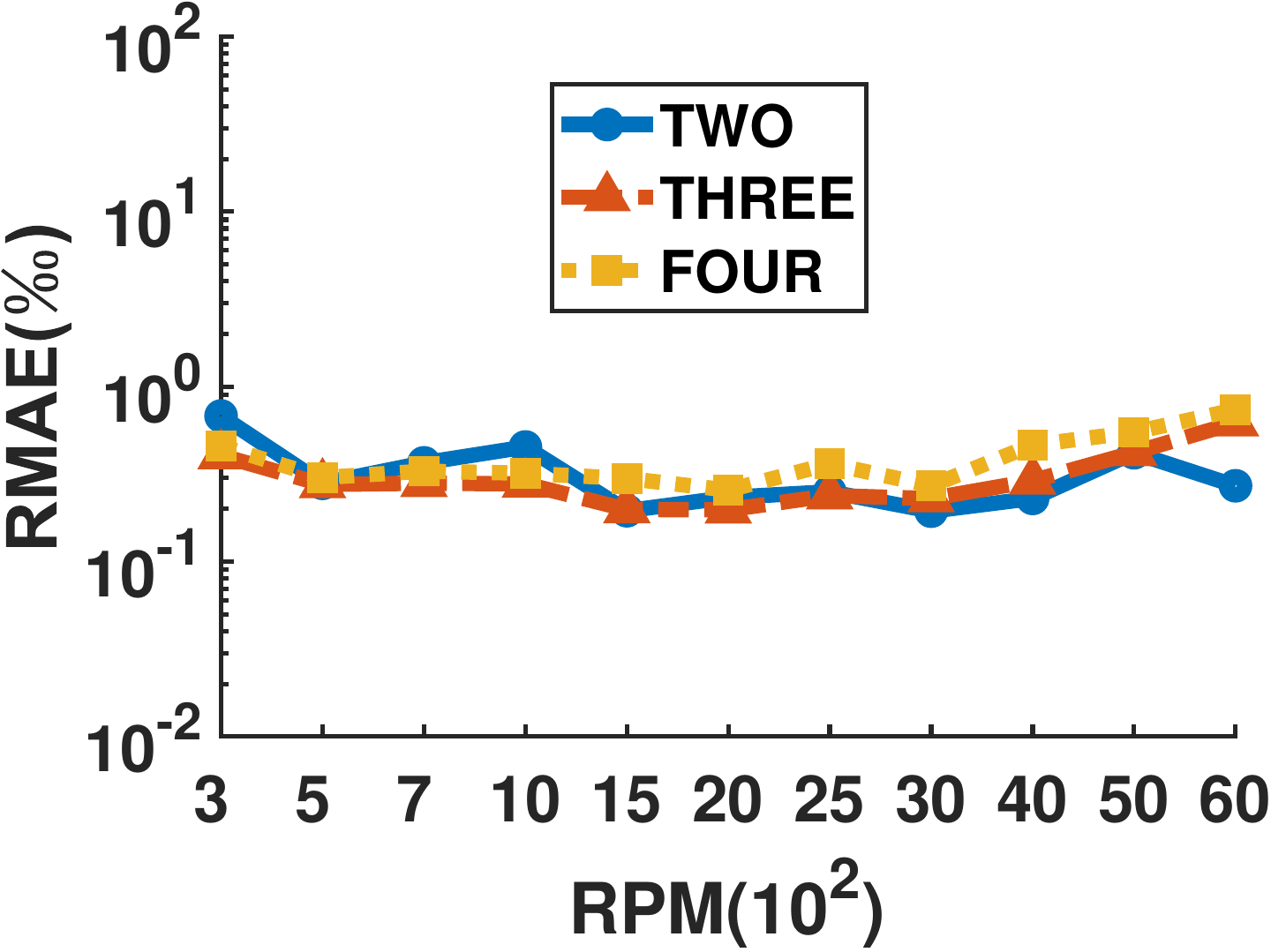}
		\caption{RMAE of different measurement methods on different number of blades}
		\label{fig:blades}

\end{figure}

\subsubsection{\bf \textit{Evaluation on Robustness to Host Vibration}}
\label{sec:evaluation_vibration}
\begin{figure}[h]
		\includegraphics[width=0.8\columnwidth]{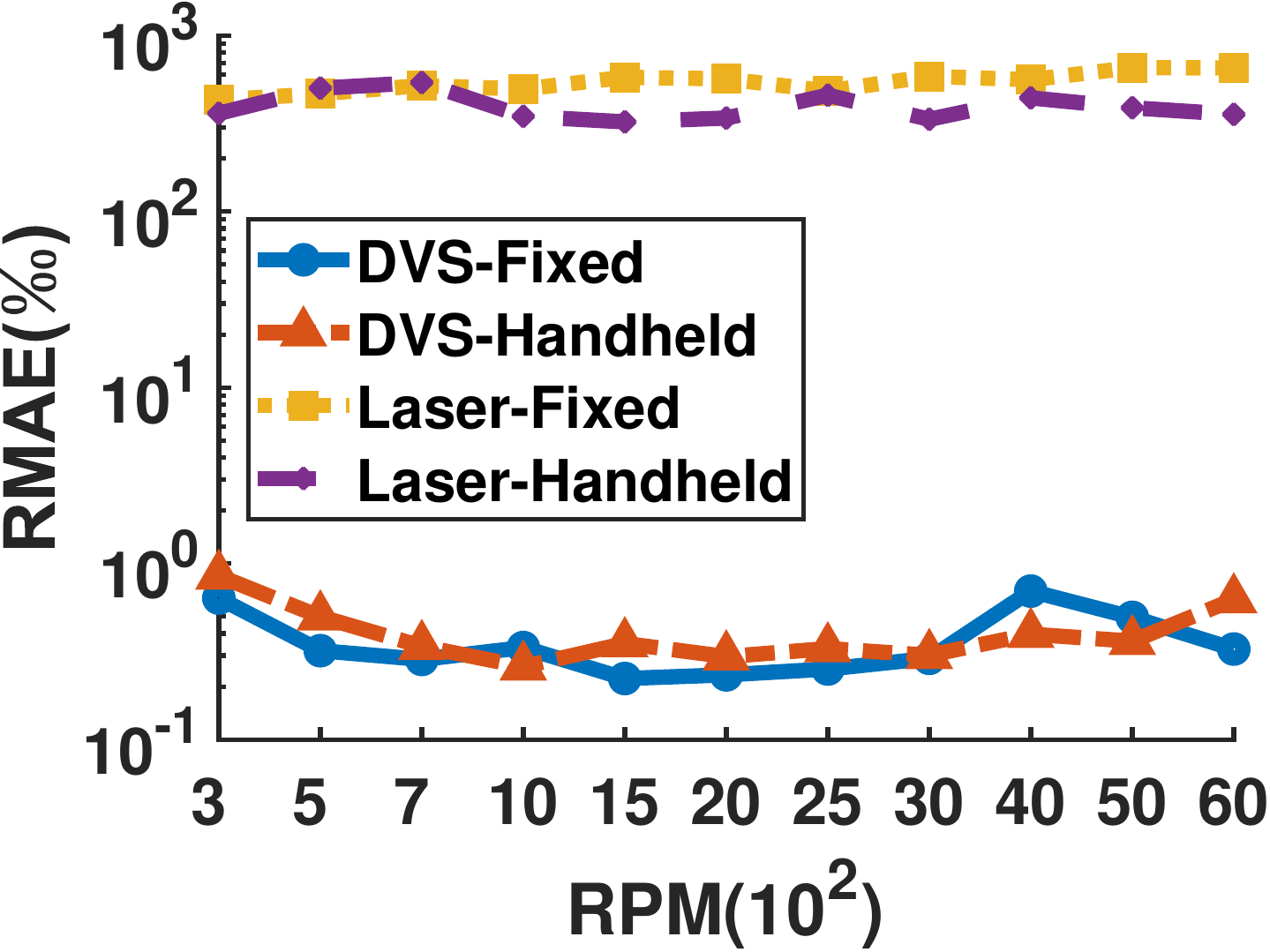}
		\caption{RMAE of different measurement methods on vibrating host.}
	   \label{fig:host}
		\end{figure}
In real-world scenarios, the hosting device are sometimes not stable, e.g., vibrating. To obtain accurate measurement, the tachometer should be able to accommodate the slight motion of the hosting device to some extent. Therefore, we simulate the vibrating host to compare the robustness of {\sl EV-Tach} and laser tachometer.
DAVIS346 and laser tachometer are placed $40$cm away from the servo motor and the servo motor is located on a table. A number of people are asked to shake the table to simulate the vibration. By gradually changing the rotational speed from $300$rpm to $6000$rpm,  the RMAEs are calculated and shown in Fig.~\ref{fig:host}. From the results we can observe, 
{\bf Laser-Fixed} and {\bf Laser-Handheld} produce significantly high RMAEs for measuring the rotating target on a vibrating host: the average of the RMAEs are $\text{547\textperthousand}$ and $\text{399\textperthousand}$ respectively which indicates laser tachometer completely fails in providing reasonable measurement results. On the contrary, {\sl EV-Tach} demonstrates great robustness to the vibrating host: the average of RMAEs is only $0.37\text{\textperthousand}$ for fixed and $0.41\text{\textperthousand}$ for handheld which is close to the stable host scenario and is almost $1500$ times better than laser tachometer.

% %==========================================
% Conclusion
% %==========================================
\section{DISCUSSION}
\label{sec:discussion}
% \noindent{\bf Advantages and Limitations:}
As the description and evaluation in this paper, {\sl EV-Tach } shows a number of superior characteristics on the task of rotational speed measurement over the state-of-the-art laser tachometer which is dominant in the market. First and foremost, {\sl EV-Tach} is a real handheld tachometer and it is robust to subtle movement of user's hand. While, though the laser tachometer is designed as a portable device, its accuracy drops significantly when used as handheld. Second, the use of {\sl EV-Tach} is more convenient than laser tachometer and no preparation is needed before measurement; while laser tachometer must be pointed to the small reflective labels and it takes users about tens of seconds when the target is fast rotating. Third, {\sl EV-Tach} is robust to the vibrating host of the rotating target which causes failed measurement for laser tachometer. Fourth, the {\sl EV-Tach} is able to measure multiple rotating targets at the same time, which is impossible for laser tachometer. Finally, compared with the other vision-based method, it achieves significantly higher range of measurement than those with conventional RGB cameras and is more cost-effective than those applying high-speed cameras. 

However, as the principal of DVS, {\sl EV-Tach} also shares some similar limitations to the vision-based approaches. First of all, it requires the rotating targets in form of propellers or with uneven texture so that different phases of rotation can be detected. However, like the reflective label for laser tachometer, the usability of {\sl EV-Tach} can be extended if unique pattern or labels are allowed to be attached on the rotating object. For example, when the rotating object is a flat disc with uniform texture, a label (e.g., a straight line), which is high contrast to the disc, can be attached to aid the measurement. Second, constrained by the hardware design, the accuracy and range of measurement of {\sl EV-Tach} in this paper is lower than laser tachometer. However considering the spatial resolution of DAVIS346 is only $346\times240$, the performance of {\sl EV-Tach} is expected to be improved by using event-cameras with higher spatial and temporal resolution.  

\section{Conclusion}
\label{sec:conclusion}
In this paper, we propose, {\sl EV-Tach}, a rotational speed measurement system based on dynamic vision sensing on mobile devices to achieve efficient high-fidelity and convenient estimation. {\sl EV-Tach} starts with extracting multiple rotating targets via K-means clustering and a heatmap-based initial centroids selection method is propose to improve the robustness of the clustering. Then angle of rotation is estimated via a coarse-to-fine ICP-based event streams registration method and rotational speed can be calculated afterwards. At last, Extensive evaluations are conducted and the results show that the accuracy of {\sl EV-Tach} is comparable to laser tachometer in fixed deployment and is over $210$ better in handheld measurement mode. {\sl EV-Tach} is robust to shaking host of rotating target in which the laser tachometer fails to provide reasonable results.

% \endinput
\bibliographystyle{ACM-Reference-Format}

\bibliography{main}

\end{document}